\documentclass[runningheads]{llncs}
\usepackage[T1]{fontenc}
\usepackage{graphicx}
\usepackage{booktabs}
\usepackage[hidelinks]{hyperref}
\usepackage{amssymb}
\usepackage{xcolor}
\usepackage{xspace}
\usepackage{tikz}
\usetikzlibrary{positioning, fit, backgrounds, arrows.meta, calc}
\newcommand{\vigil}{\textsc{Vigil}\xspace}
\begin{document}

\title{\vigil: An Extensible System for Real-Time\\Detection and Mitigation of Cognitive Bias Triggers}
\titlerunning{Vigil: Real-Time Cognitive Bias Trigger Detection}

\author{Bo Kang
\and
Sander Noels
\and
Tijl {De Bie}
}
\authorrunning{B. Kang et al.}
\institute{Ghent University, Belgium
\email{\{firstname.lastname\}@ugent.be}}

\maketitle
%
\begin{abstract}
The rise of generative AI is posing increasing risks to online information integrity and civic discourse.
Most concretely, such risks can materialise in the form of mis- and disinformation.
As a mitigation, media-literacy and transparency tools have been developed to address factuality of information and the reliability and ideological leaning of information sources~\cite{newsguard2024,hassan2017claimbuster,lees2022new}.
However, a subtler but possibly no less harmful threat to civic discourse is the use of persuasion or manipulation by exploiting human cognitive biases and related cognitive limitations~\cite{kahneman2011thinking}.
To the best of our knowledge, no tools exist to directly detect and mitigate the presence of triggers of such cognitive biases in online information.

We present \vigil (VIrtual GuardIan angeL), the first browser extension for real-time cognitive bias trigger detection and mitigation, providing in-situ scroll-synced detection, LLM-powered reformulation with full reversibility, and privacy-tiered inference from fully offline to cloud.
\vigil is built to be extensible with third-party plugins, with several plugins that are rigorously validated against NLP benchmarks already included.
It is open-sourced at \url{https://github.com/aida-ugent/vigil}.
\end{abstract}
%
%
\section{Introduction}\label{sec:intro}

Misinformation is a commonly cited threat to rational public discourse.
Also highly corrosive, however, is the rhetorical exploitation of cognitive biases in how information, whether factual or not, is \emph{presented}.
Dual-process theory~\cite{kahneman2011thinking} distinguishes fast, heuristic-driven cognition from slow, deliberate reasoning.
The fast system is prone to systematic biases that online content routinely exploits through what we term \emph{cognitive bias triggers}: textual patterns that activate specific biases in readers.
For example, loaded language triggers the affect heuristic; appeals to authority exploit deference to expertise; repetition triggers the illusory truth effect.
Readers therefore benefit not only from knowing whether information is true, but from understanding \emph{how} it is framed to affect their reasoning~\cite{van2023foolproof}.

Existing media-literacy tools operate on two dimensions.
\emph{Ideology} tools such as NewsGuard~\cite{newsguard2024}, Media Bias/Fact Check~\cite{mbfc2024}, and Ground News~\cite{groundnews2024} rate sources by political orientation and credibility, focusing on \emph{who} is the author of the information.
However, a credible source can still employ manipulative rhetoric, and a biased source can make a valid rational argument.
\emph{Factuality} tools such as ClaimBuster~\cite{hassan2017claimbuster} and Perspective API~\cite{lees2022new} identify check-worthy claims or score toxicity, assessing \emph{what} is being said.
Yet even factually correct statements can be manipulative or misleading.
Neither type of tool addresses \emph{how} the information is presented, and specifically whether the presentation contains triggers that activate \emph{cognitive biases or related limitations} in the reader, causing them to process the information irrationally.
Academic NLP research on fallacy detection~\cite{jin2022logical,lim2024evaluation}, propaganda technique detection~\cite{da2020semeval}, and moralization analysis~\cite{becker2025moralization} has produced relevant methods, but these remain offline batch tools, not embedded in the browsing experience.

{\bf Contributions.} We present \vigil, to our knowledge the first browser extension for real-time cognitive bias trigger detection and mitigation. It features:
(1)~\emph{in-situ detection and mitigation}: scroll-synced, span-level detection of cognitive bias triggers directly within the browsing experience, with LLM-powered reformulation and full reversibility;
(2)~a \emph{privacy-tiered architecture} from fully offline to cloud, letting users choose their privacy/capability trade-off;
(3)~an \emph{extensible plugin system} where new trigger types require only a new plugin, not architectural changes;
and (4)~two \emph{validated plugins} for two persuasion strategies (corresponding to a taxonomy of cognitive bias triggers, and a set of moralization strategies), both rigorously evaluated against published benchmarks.
\vigil has been designed to work particularly well on Twitter/X-feeds and news websites.

Section~\ref{sec:arch} describes the system architecture. Section~\ref{sec:demo} presents the user experience and evaluation. Section~\ref{sec:discussion} discusses related systems, responsible AI considerations, and future work.
%
%
\section{System Architecture}\label{sec:arch}
%
%
\begin{figure}[t]
  \centering
  \begin{tikzpicture}[
    ctx/.style={draw, rounded corners=1.5pt, fill=black!5,
      minimum height=9mm, minimum width=22mm, align=center, font=\small},
    plug/.style={draw, rounded corners=1.5pt, fill=black!9,
      minimum height=9mm, align=center, font=\small},
    ext/.style={draw, rounded corners=1.5pt, fill=black!5,
      minimum height=9mm, minimum width=28mm, align=center, font=\small},
    lbl/.style={font=\scriptsize\itshape, text=black!50},
    arr/.style={<->, >={Stealth[length=3.5pt]}, thin},
    sarr/.style={->, >={Stealth[length=3.5pt]}, thin},
  ]
  \node[ctx] (cl) at (0, 0) {Content Layer};
  \node[ctx] (mr) at (3.2, 0) {Message Router};
  \node[ctx] (sp) at (6.4, 0) {Sidepanel};
  \node[ctx, minimum width=24mm] (ir) at (3.2, -1.6) {Inference Runtime};
  \node[font=\tiny, text=black!40, below=0.3mm of cl] {viewport $\cdot$ render};
  \node[font=\tiny, text=black!40, below=0.3mm of mr] {routing $\cdot$ lifecycle};
  \node[font=\tiny, text=black!40, below=0.3mm of sp] {Analyze $|$ Settings};
  \node[font=\tiny, text=black!40, below=0.3mm of ir] {WebGPU / WebLLM};
  \draw[arr] (cl) -- node[above, lbl] {msgs} (mr);
  \draw[arr] (mr) -- node[above, lbl] {msgs} (sp);
  \draw[arr] (mr) -- node[right, lbl, pos=0.75] {msgs} (ir);
  %
  \node[plug, minimum width=30mm] (bp) at (0, -2.8)
    {Browser Plugins\\[-2pt]{\scriptsize\texttt{cbt-regex} $\cdot$ \texttt{cbt-llm}}};
  \node[plug, minimum width=30mm] (spl) at (6.4, -2.8)
    {Server Plugins\\[-2pt]{\scriptsize\texttt{moralization-llm}}};
  \draw[dash dot, thin] (bp.east) -- node[above, lbl] {\texttt{Finding} contract} (spl.west);
  \draw[sarr] (ir.south west) -- (bp);
  \draw[sarr] (ir.south east) -- (spl);
  %
  \begin{scope}[on background layer]
  \node[draw, thick, rounded corners=3pt,
    inner xsep=3mm, inner ysep=3mm,
    fit=(cl)(ir)(bp)(spl),
    label={[font=\small\bfseries]above:Chrome Extension}
  ] (extbox) {};
  \end{scope}
  %
  \draw[densely dashed, semithick, black!40]
    ([yshift=-1.5mm]extbox.south west) --
    node[below=0.5mm, font=\scriptsize\bfseries, text=black!40]
      {browser boundary}
    ([yshift=-1.5mm]extbox.south east);
  %
  \node[ext] (server) at (6.4, -4.7)
    {Backend Server\\[-2pt]{\scriptsize FastAPI $\cdot$ plugin registry}};
  \draw[sarr] (spl) -- node[right, lbl, pos=0.75] {HTTP} (server);
  \end{tikzpicture}
  \caption{\vigil architecture. Four extension components communicate via typed messages: the content layer extracts and renders page text, the message router coordinates, the sidepanel exposes the UI, and the inference runtime runs the in-browser LLM. Detection is delegated to browser plugins (in-process) or server plugins (HTTP), both producing \texttt{Finding} objects via a shared contract (dash-dotted line). The dashed line marks the browser boundary; components above it keep all data in-process.}\label{fig:arch}
\end{figure}
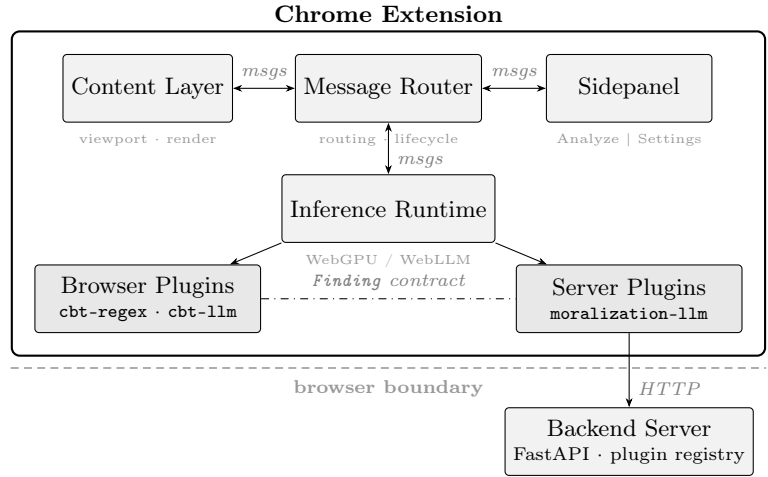
\vigil is implemented as a Chrome extension with an optional Python backend server (Fig.~\ref{fig:arch}).
The architecture separates concerns across four isolated components and supports four inference tiers with different privacy guarantees.

\subsection{Extension Architecture}\label{sec:ext-arch}

The extension comprises four components communicating via a typed message protocol:
\begin{enumerate}
  \item \textbf{Content layer}: injected into each web page, it extracts text using a dedicated Twitter/X parser or a generic page extractor, tracks which content blocks are currently in the user's viewport, and renders in-page highlights and reformulations.
  \item \textbf{Message router}: a background process that coordinates communication between components and manages the inference runtime. It holds no persistent state and is designed to be fully recoverable, as the browser may terminate it at any time.
  \item \textbf{Sidepanel}: the primary user interface, with two tabs: \emph{Analyze} (scroll-synced ``Currently Viewing'' card, findings list, action buttons) and \emph{Settings} (sensitivity, plugin selection, LLM backend, auto-analyze mode).
  \item \textbf{Inference runtime}: a dedicated process for in-browser LLM inference via WebGPU, isolated from the main extension to avoid blocking the UI.
\end{enumerate}
A key design choice is \emph{viewport tracking}: the content layer monitors which content blocks are visible as the user scrolls, driving the sidepanel's ``Currently Viewing'' card.
This enables scroll-synced analysis without manual content selection; the system always knows what the user is looking at.

\subsection{Plugin System}\label{sec:plugins}

All detection logic is encapsulated in plugins that share a common \texttt{Finding} contract (trigger type, severity, text span, explanation), enabling uniform display regardless of detection method.
Two plugin types are supported:

\paragraph{Browser plugins} run in-process within the extension.
Two are included. The first, \texttt{cbt-regex} (cognitive bias trigger -- regex), performs keyword and phrase pattern matching against a 14-type cognitive bias taxonomy, with zero latency and no LLM required, serving as an always-available fallback.
The second, \texttt{cbt-llm} (cognitive bias trigger -- LLM), detects the same 14 cognitive bias types from the same taxonomy, but does this via LLM inference.
The taxonomy is derived from the SemEval-2020 Task~11 on the detection of propaganda techniques~\cite{da2020semeval}, from which we mapped each propaganda technique to the cognitive bias it exploits (e.g., Loaded Language $\to$ affect heuristic, Appeal to Authority $\to$ authority bias, Repetition $\to$ illusory truth effect).
Each finding includes the trigger label, the cognitive bias being triggered, the severity, and an explanation.
Both browser plugins share a single-source-of-truth taxonomy JSON file used across detection, evaluation, and UI display.

\paragraph{Server plugins} run on an optional Python backend and are discovered at runtime via a plugin registry.
The server plugin included in this paper, \texttt{moralization-llm}, performs moralization detection grounded in Moral Foundations Theory~\cite{haidt2012righteous} and the Moralization Corpus annotation scheme~\cite{becker2025moralization}, identifying moral values across 12 categories, explicit or implicit demands, and protagonist roles. It supports both English and German.

Adding a new trigger type requires only implementing the plugin interface; no changes to the extension architecture or UI are needed.

\begin{table}[h]
  \caption{Privacy tiers. Each backend offers a different privacy--capability trade-off. Tiers~1--2 are \emph{verifiably} zero-network: no data leaves the machine.}\label{tab:privacy}
  \centering
  \begin{tabular}{@{}llcc@{}}
    \toprule
    Tier & Backend & Leaves browser & Leaves machine \\
    \midrule
    Regex        & In-browser patterns         & No  & No  \\
    WebGPU LLM   & In-browser (e.g., Llama 3.2 1B)   & No  & No  \\
    Local API    & Local server (user's model)   & Yes & No  \\
    Cloud API    & Remote server (e.g.\ OpenAI) & Yes & Yes \\
    \bottomrule
  \end{tabular}
\end{table}

\subsection{Privacy-Tiered Inference}\label{sec:privacy}

Table~\ref{tab:privacy} summarizes the four inference tiers, ranging from fully offline to cloud.
The WebGPU tier uses in-browser LLM inference via WebLLM~\cite{ruan2024webllm}; text never leaves the browser, and this guarantee is \emph{verifiable} (privacy-by-design).
Regex provides instant screening, while LLM tiers add depth on demand.
No data is persisted server-side in any tier, and results are cached locally with automatic eviction.
%
%
\section{User Experience and Evaluation}\label{sec:demo}

We now discuss the \vigil user-experience, as well as an evaluation of the provided plugins.

\subsection{User Experience}\label{sec:walkthrough}
%
\begin{figure}[t]
  \centering
  \includegraphics[width=\textwidth]{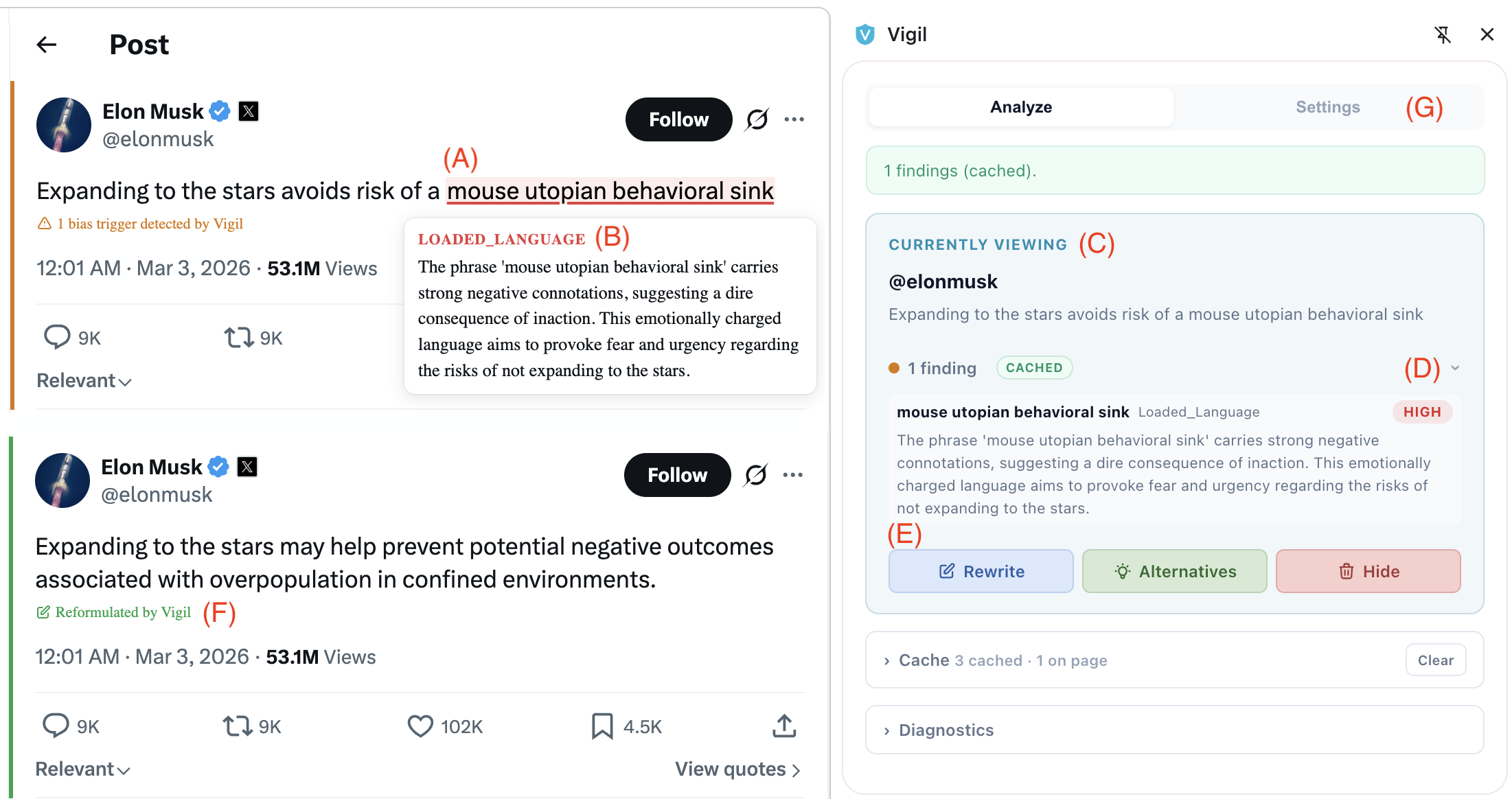}
  \caption{\vigil analyzing a tweet on Twitter/X.
  (A)~A detected cognitive bias trigger is underlined in-page with a severity-colored highlight.
  (B)~Hovering reveals the bias type (\textsc{Loaded Language}) and an explanation.
  (C)~The sidepanel's scroll-synced ``Currently Viewing'' card tracks the post in the user's viewport.
  (D)~The finding card details the trigger span, bias type, severity, and explanation.
  (E)~Action buttons let the user rewrite the text to neutralize framing, view alternatives, or hide the post.
  (F)~After reformulation, neutralized but semantically equivalent text replaces the original in-place. The Rewrite button becomes ``Restore Original'' for full reversibility.
  (G)~The Settings tab provides plugin selection and LLM backend configuration, controlling the privacy--capability trade-off (Table~\ref{tab:privacy}).}\label{fig:screenshots}
\end{figure}

Typical use of the \vigil extension consists of the following steps (Fig.~\ref{fig:screenshots}).\footnote{Video walkthrough: \url{https://aida.ugent.be/videos/vigil-v001-demo.mp4}}
\begin{itemize}
  \item[(1)] \textbf{Browse.} The user opens Twitter/X or a pre-loaded news page with \vigil installed. The sidepanel opens alongside the page, showing the analysis interface.
  \item[(2)] \textbf{Scroll.} As the user scrolls, the ``Currently Viewing'' card~(C) in the sidepanel tracks the primary visible content item in real time (driven by viewport tracking). No manual selection needed.
  \item[(3)] \textbf{Analyze.} Clicking ``Analyze'' triggers the analysis pipeline, after which the detected text span is highlighted directly in the page~(A) with a hover tooltip~(B), and the sidepanel shows a finding card with severity badge, bias type, and explanation~(D).
  \item[(4)] \textbf{Mitigate.} The user clicks ``Rewrite''~(E) on a finding. The LLM reformulates the text; neutralized but semantically equivalent text replaces the original on-page with a visual indicator~(F). The Rewrite button becomes ``Restore Original'' for full reversibility.
  \item[(5)] \textbf{Explore.} The user can also request ``Alternatives'' or ``Hide''~(E), shown in the sidepanel.
  \item[(6)] \textbf{Configure.} The Settings tab~(G) exposes plugin selection and the privacy-tiered backend choice (Table~\ref{tab:privacy}).
\end{itemize}

\paragraph{Auto-analyze mode.} When the auto-analyze mode is enabled via the Settings tab (Fig.~\ref{fig:screenshots}G) with a default action (rewrite or hide), every newly visible content item is immediately obscured while the LLM is working, so the user never sees potentially manipulative text before the action is applied.

%
\subsection{Evaluation}\label{sec:eval}
We not only present the software framework, but also include high-quality plugins with state-of-the-art detection quality on established benchmarks.
We evaluate both detection quality and system latency.

\paragraph{Cognitive bias trigger detection.}
We evaluate the \texttt{cbt-llm} plugin on SemEval-2020 Task~11~\cite{da2020semeval}, using the protocol from Sprenkamp et al.~\cite{sprenkamp2023large}: span-level propaganda technique annotations are aggregated to per-article multi-label sets (14 propaganda technique types, 75-article development split), and micro-F1 is computed over these label sets.
Since, to the best of our knowledge, no dedicated cognitive bias trigger benchmark exists, we use this dataset as a proxy: each propaganda technique maps onto a specific cognitive bias it exploits (Section~\ref{sec:plugins}), and the LLM-based results from Sprenkamp et al.\ serve as our validation baseline.
Using GPT-4o (to match the model used by Sprenkamp et al.) with the same base prompt, we obtain micro-F1\,=\,0.581, matching their reported GPT-4 result within 0.02 points.
The plugin's production prompt, which requires structured output with text spans, explanations, and cognitive bias annotations, achieves a very competitive micro-F1\,=\,0.533 with precision\,=\,0.626, deliberately favoring precision.
We argue that for a user-facing tool, this trade-off makes sense: false alarms erode trust more than missed detections.

\paragraph{Moralization detection.}
We evaluate the \texttt{moralization-llm} plugin on the Test-150 split of the Moralization Corpus~\cite{becker2025moralization}, a 150-instance binary detection benchmark (German texts, 32\,\% positive rate) annotated by three experts and two students across parliamentary debates, news commentary, letters, court reports, non-fiction, and online discussions.
Macro F1 is the benchmark's primary metric.
Using gpt-4o-mini (the production backend; Table~\ref{tab:latency}), the plugin achieves macro F1\,=\,0.789 (German prompt) and 0.753 (English prompt), competitive with the corpus authors' best result (Cohere, macro F1\,=\,0.772).
Model--human agreement, measured by PABAK~\cite{byrt1993bias} (prevalence- and bias-adjusted kappa) against the five annotators, reaches 0.582, approaching expert--expert agreement (0.698).

\paragraph{System latency.}

\begin{table}[t]
  \caption{Analysis of the latency across privacy tiers (Apple M3 Pro, 36\,GB RAM; 15-text corpus across short/medium/long bins). WebGPU cold start (one-time model load): ${\sim}$21\,s.}\label{tab:latency}
  \centering
  \begin{tabular}{@{}llrrrc@{}}
    \toprule
    Tier & Backend & Median & P95 & $N$ & Network \\
    \midrule
    Regex    & In-browser patterns       & 0.03\,ms & 0.10\,ms & 750 & None \\
    WebGPU   & In-browser (Llama 3.2 1B) & 3.4\,s   & 15.9\,s  & 45  & None \\
    Server   & Cloud (gpt-4o-mini)       & 3.9\,s   & 9.1\,s   & 300 & HTTP \\
    \bottomrule
  \end{tabular}
\end{table}

Table~\ref{tab:latency} confirms that analysis stays within an interactive range across the benchmarked tiers.
Regex is sub-millisecond, substantiating the real-time claim for pattern-based detection.
The two LLM tiers converge to similar medians (WebGPU 3.4\,s, Server 3.9\,s), meaning on-device processing matches cloud speed in the typical case, and the privacy tier imposes no latency penalty.
WebGPU's higher P95 (15.9\,s) reflects long input texts rather than a systemic cost. Caching makes repeated views instantaneous regardless of tier.
%
%
\section{Discussion}\label{sec:discussion}

Next, we compare \vigil with comparable existing tools, discuss some responsible AI considerations that had an impact on its design, and end with some conclusions and suggestions for further work.

\subsection{Comparison with Existing Systems}\label{sec:comparison}

Table~\ref{tab:comparison} compares \vigil against existing tools by dimension, granularity, and functionalities.
In the ideology dimension, tools range from site-level ratings (NewsGuard~\cite{newsguard2024}, MBFC~\cite{mbfc2024}) through article-level bias distribution (Ground News~\cite{groundnews2024}) to per-tweet credibility nudges (NudgeCred~\cite{bhuiyan2021nudgecred}).
All operate in-situ as browser extensions, but none perform real-time NLP analysis or offer content-level mitigation; they rely on pre-curated databases of source ratings.
In the factuality dimension, Perspective API~\cite{lees2022new} provides real-time toxicity scoring but as a developer API, not a user-facing tool.
ClaimBuster~\cite{hassan2017claimbuster} identifies check-worthy claims but is not embedded in the browsing experience.

\begin{table}[t]
    \caption{Comparison with other media-literacy tools. \vigil is the first to combine in-situ delivery, real-time detection, and mitigation in the cognitive processing dimension.}\label{tab:comparison}
    \centering
    \small
    \begin{tabular}{@{}lccccc@{}}
      \toprule
      System & Dimension & Granularity & In-situ & Real-time & Mitigation \\
      \midrule
      NewsGuard~\cite{newsguard2024}       & Ideology    & Site    & \checkmark & --         & -- \\
      MBFC~\cite{mbfc2024}                 & Ideology    & Site    & \checkmark & --         & -- \\
      Ground News~\cite{groundnews2024}    & Ideology    & Article & \checkmark & --         & -- \\
      NudgeCred~\cite{bhuiyan2021nudgecred}& Ideology    & Post    & \checkmark & --         & -- \\
      Perspective~\cite{lees2022new}       & Factuality  & Post    & --         & \checkmark & -- \\
      ClaimBuster~\cite{hassan2017claimbuster} & Factuality & Sentence & --         & \checkmark & -- \\
      Prta~\cite{da2020prta}               & Cognitive   & Span    & --         & \checkmark & -- \\
      \midrule
      \textbf{\vigil}   & \textbf{Cognitive} & \textbf{Span} & \checkmark & \checkmark & \checkmark \\
      \bottomrule
    \end{tabular}
  \end{table}

The cognitive processing dimension is largely unoccupied.
The closest system is Prta~\cite{da2020prta}, which performs span-level propaganda detection but as a standalone web application without in-situ integration or mitigation; its published demo URL currently returns 404 (checked March 2026).
Academic work on fallacy detection~\cite{jin2022logical,lim2024evaluation} has produced classifiers, but these remain offline batch tools.
\vigil is, to our knowledge, the first browser extension delivering real-time in-situ detection and mitigation in the cognitive processing dimension.

It is worth distinguishing \vigil from toxicity detection: Perspective API measures how offensive content is, while \vigil is designed to detect \emph{cognitive bias triggers} and related rhetorical devices that may be present in non-toxic, factually correct text.
In dual-process terms~\cite{kahneman2011thinking}, ideology and factuality tools help users judge truth value, a deliberate, ``slow system'' task.
\vigil addresses a prior step: recognizing when the ``fast system'' is being targeted by rhetorical strategies, so the user can engage deliberate reasoning before judging truth value.

\subsection{Responsible AI Considerations}\label{sec:responsible}

Several responsible AI considerations informed the design of \vigil. We discuss these below, as well as several limitations of the current implementation.

\paragraph{Privacy-by-design.}
Two of four inference tiers (regex, WebGPU) are verifiably zero-network: no data leaves the browser.
No telemetry or usage data is collected in any tier. Users choose their data exposure through the settings panel.

\paragraph{No censorship.}
\vigil surfaces patterns and offers tools; it never blocks or removes content unilaterally.
All interventions are reversible with one click.
The system is designed to \emph{augment} the user's judgment, not replace it.

\paragraph{Transparency and grounding.}
\vigil is open-source. Each finding includes an explanation and confidence score.
Plugins are built on published taxonomies (SemEval-2020 Task~11~\cite{da2020semeval} propaganda techniques mapped to cognitive biases for trigger detection, Moral Foundations Theory~\cite{haidt2012righteous} for moralization), not ad-hoc heuristics.

\paragraph{Current limitations.}
LLM hallucination can produce false positives, mitigated by human-in-the-loop design (all interventions require user action) but not eliminated.
The SemEval propaganda corpus serves as a proxy for cognitive bias triggers; to the best of our knowledge, a dedicated benchmark with cognitive-science grounding does not yet exist (Section~\ref{sec:conclusion}), and likewise we are unaware of any equivalent benchmark for evaluating mitigation quality.
Coverage is currently limited to English and German, with platform support for Twitter/X (dedicated parser) and generic web pages.

\subsection{Conclusion and Future Work}\label{sec:conclusion}

We presented \vigil, to our knowledge the first browser extension for real-time detection and mitigation of cognitive bias triggers and related rhetorical techniques, a third dimension to media-literacy tools orthogonal to ideology and factuality.
Grounded in dual-process theory and published NLP benchmarks, \vigil provides real-time detection of rhetorical patterns that exploit cognitive biases, with user-controlled privacy guarantees.

A key open challenge is creating a dedicated cognitive bias trigger benchmark with annotations grounded in cognitive science, linking textual triggers to the specific biases they exploit and moving beyond the current propaganda-as-proxy evaluation.
Other future directions include an evaluation framework for mitigation quality (assessing whether rewrites preserve meaning while neutralizing cognitive bias triggers), additional platform parsers and multi-language support, a personal bias exposure dashboard that tracks which cognitive biases a user encounters over time, and user studies on how trigger-aware information consumption affects rational information consumption, belief formation, and decision making.
\paragraph{Acknowledgements.} The research leading to these results was co-funded by the European Union (ERC, VIGILIA, 101142229), the Special Research Fund (BOF) of Ghent University (BOF20/IBF/117), the Flemish Government under the ``Onderzoeksprogramma Artificiële Intelligentie (AI) Vlaanderen'' programme, and the FWO (project no. G073924N). Views and opinions expressed are however those of the author(s) only and do not necessarily reflect those of the European Union or the European Research Council Executive Agency. Neither the European Union nor the granting authority can be held responsible for them. For the purpose of Open Access the author has applied a CC BY public copyright license to any Author Accepted Manuscript version arising from this submission.
%
%
%
%
\bibliographystyle{splncs04}
\bibliography{refs}
\end{document}